%% file: tackling_climate_workshop.tex
\documentclass{article}

     \PassOptionsToPackage{numbers, compress}{natbib}

\usepackage[final]{tackling_climate_workshop_style}

\usepackage[utf8]{inputenc} 
\usepackage[T1]{fontenc}    
\usepackage{hyperref}       
\usepackage{url}            
\usepackage{booktabs}       
\usepackage{amsfonts}       
\usepackage{nicefrac}       
\usepackage{microtype}      
\usepackage{graphicx}
\usepackage{amsmath}

\title{Multi-branch Spatio-Temporal Graph Neural Network For Efficient Ice Layer Thickness Prediction}

\author{%
  Zesheng Liu \\
  Department of Computer Science and Engineering\\
  Lehigh University\\
  Bethlehem, PA 18015 \\
  \texttt{zel220@lehigh.edu} \\
  \And
  Maryam Rahnemooonfar\thanks{Corresponding Author} \\
  Department of Computer Science and Engineering \\
  Department of Civil and Environmental Engineering\\
  Lehigh University
  Bethlehem, PA 18015 \\
  \texttt{maryam@lehigh.edu} \\
}

\begin{document}

\maketitle

\begin{abstract}
Understanding spatio-temporal patterns in polar ice layers is essential for tracking changes in ice sheet balance and assessing ice dynamics. While convolutional neural networks are widely used in learning ice layer patterns from raw echogram images captured by airborne snow radar sensors, noise in the echogram images prevents researchers from getting high-quality results. Instead, we focus on geometric deep learning using graph neural networks, aiming to build a spatio-temporal graph neural network that learns from thickness information of the top ice layers and predicts for deeper layers. In this paper, we developed a novel multi-branch spatio-temporal graph neural network that used the GraphSAGE framework for spatio features learning and a temporal convolution operation to capture temporal changes, enabling different branches of the network to be more specialized and focusing on a single learning task. We found that our proposed multi-branch network can consistently outperform the current fused spatio-temporal graph neural network in both accuracy and efficiency.
\end{abstract}

\input{introduction}
\input{dataset}

\input{keydesign}
\input{experiment}
\input{conclusion}

\begin{ack}
This work is supported by NSF BIGDATA awards (IIS- 1838230, IIS-1838024), IBM, and Amazon. We acknowledge data and data products from CReSIS generated with support from the University of Kansas and NASA Operation IceBridge.

\end{ack}

\bibliographystyle{plainnat}
\bibliography{bib}
\end{document}

%% file: introduction.tex
\section{Introduction}

As global temperatures continue to rise, research has shown that the accelerated mass loss of polar ice sheets is increasingly contributing to climate change\cite{Forsberg2017, Mouginot2019,Rignot2011, Zwally2011}. Polar ice sheets comprise several internal ice layers formed in different years. A better understanding of the status of internal ice layers can provide valuable information on snowfall melting and accumulation and enable a comprehensive understanding of the global climate system and future climate change.

Traditional methods to study the internal ice layer are through onsite ice core\cite{PATERSON1994378}. However, the limited and discrete coverage makes it impossible to study the continuous change of the ice layer. Additionally, onside ice cores are expensive to obtain and will cause destructive damage to the ice sheet. In recent years, airborne snow radar sensors have proven to be a more effective way to study ice layers. Internal layers with different depths are captured continuously as radargrams by measuring the reflected signal strength\cite{Arnold_2020}, shown in Figure\ref{fig:diagram}(\textbf{a}).

With the development of deep learning techniques, various convolution-based neural networks\cite{Rahnemoonfar_2021_JOG,DeepIceLayerTracking,DeepLearningOnAirborneRadar,Yari_2021_JSTAR} have been proposed to extract ice layer boundaries from radargrams. However, noise in the radargrams is shown to be a major obstacle to achieve high-quality results. Instead of the convolution-based neural networks, Zalatan et al.\cite{Zalatan2023,Zalatan_igarss,zalatan_icip} focus on learning the relationship between ice layers formed in different years and graph neural network. They utilized a fused spatio-temporal graph neural network, AGCN-LSTM, and aimed to make predictions on the thickness of deeper ice layers based on the thickness information of shallow layers. In their network, graph convolutional network(GCN) for spatio features is fused into the long short-term memory(LSTM) structure for temporal changes. Although their proposed methods have decent performance in predicting ice thickness, training the fused spatio-temporal graph neural network usually takes a few hours due to high computational costs.

In this paper, we aim to build upon the work of Zalatan et al.\cite{Zalatan2023,Zalatan_igarss,zalatan_icip} by improving the network performance in both accuracy and efficiency. Our major contributions are: 1) We developed a novel multi-branch spatio-temporal graph neural network that uses the GraphSAGE framework for spatio feature learning and uses temporal convolution operation to learn temporal changes, making different parts of the network more specialized to one single learning task. 2) We conducted extensive experiments on comparison with fused spatio-temporal graph neural networks, and results show that our proposed network can consistently outperform other methods in efficiency and accuracy. 

%% file: dataset.tex
\section{From Radargram To Graph Dataset}

\begin{figure}[b]
  \centering
  \includegraphics[width=\textwidth]{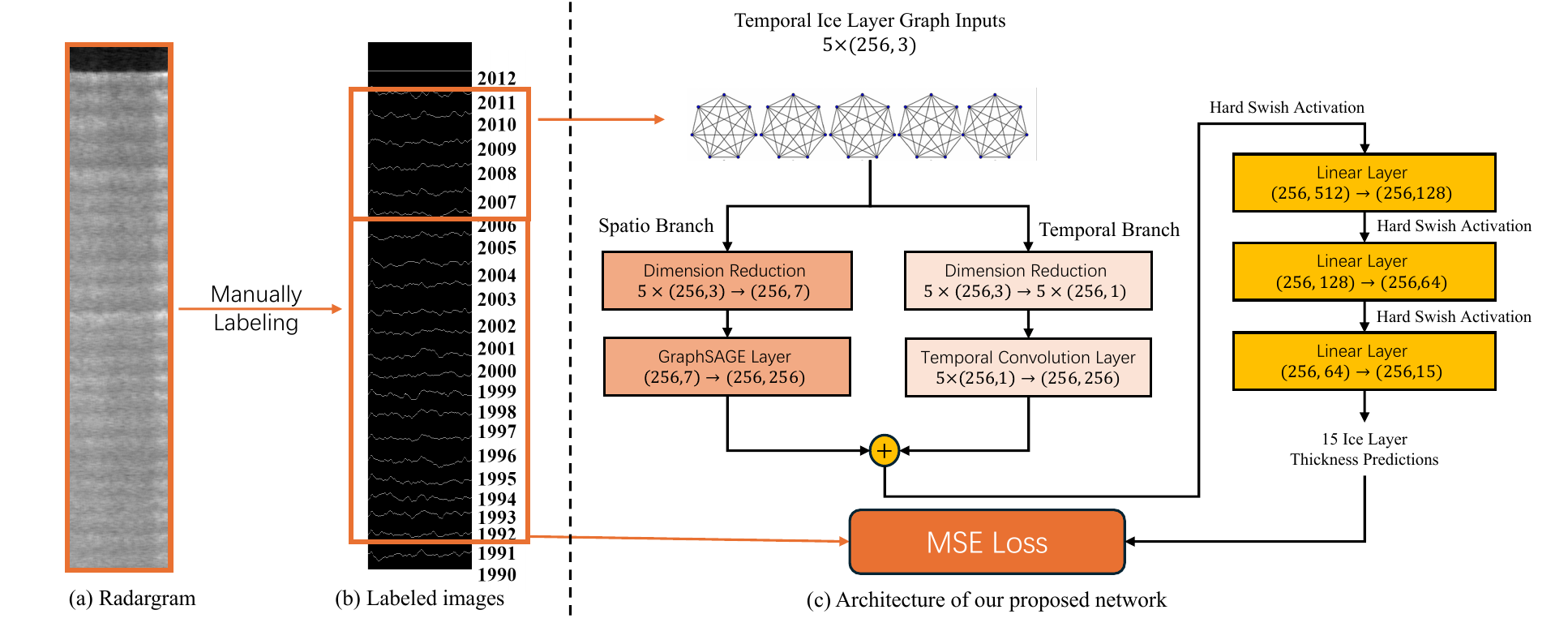}
  \caption{Diagram of our graph dataset and network structure. (\textbf{a}) Radargram captured by airborne snow radar sensor. (\textbf{b}) Labeled image where each ice layer is labeled out from the radargram. (\textbf{c}) Architecture of our proposed multi-branch network.\label{fig:diagram}}
\end{figure}

In this work, we will start with radargrams captured over the Greenland Region in 2012 via airborne snow radar sensor operated by CReSIS\cite{CReSIS_radar}. Each radargram varies from 1200 pixels to 1700 pixels in depth, with a fixed 256 pixels in width. The value of each pixel is determined by the reflected signal strength, where brighter pixels in the radargram mean a stronger reflection\cite{Arnold_2020}, shown in Figure \ref{fig:diagram}(\textbf{a}). Based on radargrams, corresponding labeled images are generated manually by NASA scientists by tracking out each layer, as illustrated in Figure Figure \ref{fig:diagram}(\textbf{b}). Based on the position of ice layer boundary pixels, layer thickness can be calculated as the difference between its upper and lower boundaries. Moreover, while capturing the radargrams, airborne snow radar sensors will also keep records of current latitude and longitude simultaneously.

Our graph dataset is then generated on the thickness and geographical location information from the radargrams and labeled images with additional pre-processing steps. As shown in Figure \ref{fig:diagram}(\textbf{b}), we will use the top five layers (formed in 2007-2011) to predict the thickness of the deeper ice layers (formed in 1992-2006). To ensure the high quality of our graph dataset, we will eliminate images with less than 20 complete layers, dividing the remaining images into training, testing, and validation sets with a ratio of $3:1:1$. This pre-processing step reduces the total number of valid images to 1660. Each valid labeled image will be represented as two temporal sequences of spatial graphs: one sequence of five graphs as input and another sequence of fifteen as the ground truth. Each spatial graph represents one internal ice layer formed in a specific year, with 256 nodes that correspond to the 256 pixels in the labeled images. Nodes will be fully connected and undirected. Edge weights between two nodes $i, j$ are defined as
\begin{equation}
    w_{i,j}  = \frac{1}{2\arcsin{(hav(\phi_j-\phi_i)+\cos{\phi_i}\cos{\phi_j}hav(\lambda_j-\lambda_i))}}
\end{equation}
where $hav(\theta) = \sin^2{(\frac{\theta}{2})}$. Each node will contain three node features: latitude, longitude, and the thickness of the current ice layer.

%% file: keydesign.tex
\section{Key Designs}
The key idea of our proposed multi-branch network is to make different parts of the network more specialized. Unlike previous AGCN-LSTM and GraphSAGE-LSTM, in which the network learns spatio and temporal features in one block, we will have individual branches to learn spatio patterns or temporal changes. Figure \ref{fig:diagram}(\textbf{c}) shows the architecture of our proposed network. For each branch, we will have a dimension reduction block to let the GraphSAGE framework or temporal convolution focus more on the relevant features. Outputs of each branch are concatenated together and passed into three linear layers for final prediction.

\subsection{Dimension Reduction}
In order to let each branch focus only on relevant features, we will have dimension reduction blocks for different branches. Based on signal transmission properties, pixels in the same pixel column but different ice layers will have the same latitude and longitude. Therefore, for the spatial branch, we will condense the overall node feature matrix of the input graph sequence from $5 \times (256, 3)$ to $(256, 7)$, where the seven node features are latitude, longitude and concatenated thickness information of the previous five ice layers. For the temporal branch, we will modify the dimension of the node feature matrix to $5 \times(256, 1)$ by removing the latitude and longitude.

\subsection{GraphSage Inductive Framework}
GraphSAGE\cite{hamilton2018GraphSAGE} is an inductive framework that generates node embedding based on a sampling and aggregating process on its local neighbors\cite{hamilton2018GraphSAGE,ZHOU202057}. For a known node $i$ with node feature $\textbf{x}_i$, it is defined as follows: 
\begin{equation}
\textbf{x}'_i = \textbf{W}_1 \textbf{x}_i + \textbf{W}_2 \cdot \text{mean}_{j \in \mathcal{N}(i)} \textbf{x}_j
\label{equation:graphsage}
\end{equation}
where $\textbf{x}'_i$ is the learned node embedding via GraphSAGE, $\textbf{W}_1, \textbf{W}_2$ are layer weights, $\mathcal{N}(i)$ is the neighbor list of node $i$ that may include neighbors with different depth, $\textbf{x}_j$ is the neighbor node features and $\text{mean}$ is the aggregator function. GraphSAGE can be understood as a linear approximation of localized spectral convolution\cite{hamilton2018GraphSAGE}, and the sampling and aggregating process can reduce the effect of possible outliers and noise and increase the model's generalization ability. 

\subsection{Temporal Convolution}
Recurrent neural networks, like long short-term memory(LSTM), can effectively learn temporal changes with high computational cost and long training time. In this work, we will replace the LSTM structure with a gated temporal convolution block proposed by Yu et al.\cite{Yu_2018} to improve its efficiency.

Inspired by Gehring et al.\cite{gehring2017convolutional}, the temporal convolution block will learn the temporal features from the original node features. An input graph sequence with node feature matrix $X$ will pass through three two-dimensional convolution operation to achieve $P, Q, R$. $P, Q$ are then passed into a gated linear unit (GLU) and $R$ is added to the output as a skip connection. Therefore, the gated temporal convolution is defined as follows:
\begin{equation}
    X' = ReLU(P\times\sigma(Q)+R)
\end{equation}
where $X'$ is the learned temporal features, $\times$ is an element-wise Hadamard product, and $\sigma$ is the sigmoid function that serves as the gate part of the GLU.

%% file: experiment.tex
\section{Experiment and Results}

To verify the design of our proposed multi-branch spatio-temporal graph neural network and the choice of branches, we compare its performance and efficiency with several different graph neural networks, including GCN-LSTM\cite{seo2016_gcnlstm} and SAGE-LSTM\cite{liu2024_sagelstm}. All the networks are trained on 8 NVIDIA A5000 GPUs and Inter(R) Xeon(R) Gold 6430 CPU. Mean-squared error loss is used as the loss function for all the networks. Adam optimizer\cite{kingma2017adam} with 0.01 as an initial learning rate and 0.0001 as a weight decay coefficient. A step learning rate scheduler halves the learning rate every 75 epochs.

\begin{figure}[b]
  \centering
  \includegraphics[width=0.6\textwidth]{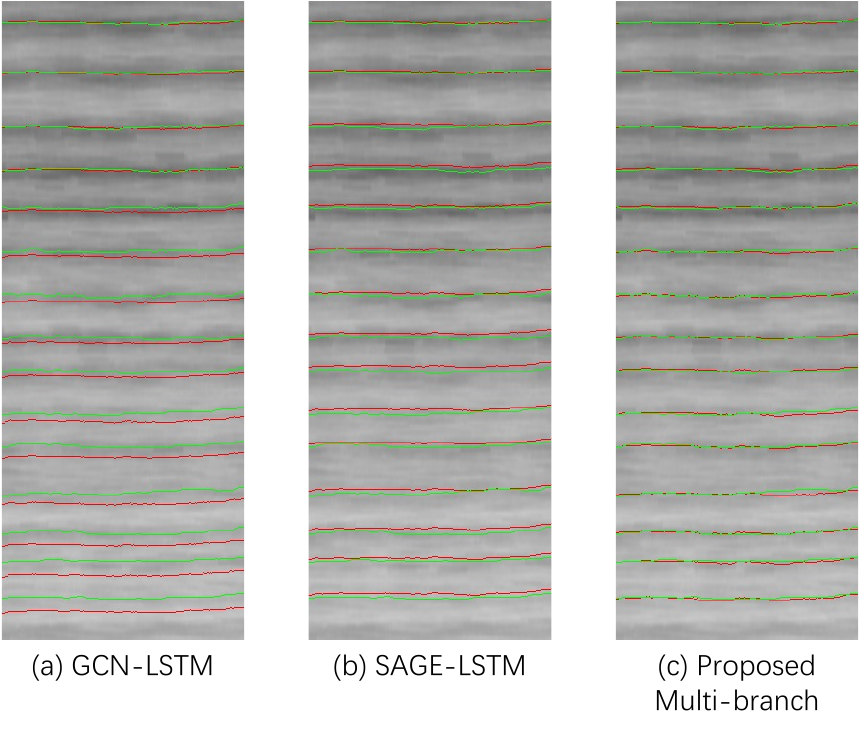}
  \caption{Qualitative results of model predictions. The green line is the groundtruth (manually-labeled ice layers) and the red line is the model prediction.\label{fig:qualitative}}
\end{figure}

The mean of the training time and the mean and standard deviations of prediction error over the five trials are reported as the model efficiency and accuracy, shown in Table \ref{table:result}. Our proposed multi-branch spatio-temporal graph neural network can significantly outperform previous fused spatio-temporal networks in both efficiency and accuracy. Figure \ref{fig:qualitative} shows the qualitative results of our trained model. Compared with the previously fused spatio-temporal graph neural networks, our proposed multi-branch method significantly improves the accuracy for bottom layers and pixels around the image boundaries.

\begin{table}
    \caption{Experiment results of GCN-LSTM, GraphSAGE-LSTM, and our proposed Multi-branch model. Results are reported as the mean and standard deviation of the RMSE on the test dataset over five individual trials. Train time is reported as the average train time over five individual trials}
    \label{table:result}
    \centering
    \begin{tabular}{ccc}
    \toprule
\textbf{Model}        & \textbf{RMSE Results} & \textbf{Average Train Time} \\ 
\midrule
GCN-LSTM              & 3.2106 $\pm$ 0.1188   & 1:58:56    \\
GraphSAGE-LSTM        & 3.1949 $\pm$ 0.0332   & 1:16:14    \\
Proposed Multi-branch(SAGE+TempConv) & \textbf{3.1236 $\pm$ 0.0548}   & \textbf{0:16:18}    \\ 
\bottomrule
\end{tabular}
\end{table}

%% file: conclusion.tex
\section{Conclusion}
In this work, we develop a novel multi-branch spatio-temporal graph neural network for learning deeper ice layer thickness. This network utilizes GraphSAGE as the spatial branch and temporal convolution to learn temporal changes over time. Experiments show that compared with previous fused spatio-temporal networks, our proposed method consistently performs better in both accuracy and efficiency.